\documentclass{article}
\usepackage[utf8]{inputenc}
\usepackage[margin=8em]{geometry}
\usepackage{graphicx}
\usepackage{amsmath}
\usepackage{amsthm}
\usepackage{enumitem}
\usepackage{booktabs}
\usepackage{makecell}

\newtheoremstyle{exampstyle}
  {0.2em} 
  {0.em} 
  {} 
  {} 
  {\bfseries} 
  {.} 
  {.5em} 
  {} 
\theoremstyle{exampstyle} \newtheorem{thm}{Theorem}
\theoremstyle{exampstyle} \newtheorem{defn}{Definition}

\newcommand{\Hull}{\text{Hull}}
\usepackage{authblk}
\usepackage{natbib}
\usepackage{amssymb}

\usepackage{bbm}

\def \bx{\boldsymbol{x}}

\def \bX{\boldsymbol{X}}

\usepackage{url}
\title{Learning in High Dimension Always Amounts to Extrapolation
}
\author{Randall~Balestriero$^1$}
\author{J\'er\^ome~Pesenti$^1$}
\author{Yann~LeCun$^{1,2}$}
\affil{$^1$Facebook~AI~Research, $^2$NYU\\{\footnotesize \{rbalestriero,pesenti,yann\}@fb.com}}
\date{}

\begin{document}

\maketitle
\begin{abstract}
    The notion of interpolation and extrapolation is fundamental in various fields from deep learning to function approximation. Interpolation occurs for a sample $\bx$ whenever this sample falls inside or on the boundary of the given dataset's convex hull. Extrapolation occurs when $\bx$ falls outside of that convex hull. One fundamental (mis)conception is that state-of-the-art algorithms work so well because of their ability to correctly interpolate training data. A second (mis)conception is that interpolation happens throughout tasks and datasets, in fact, many intuitions and theories rely on that assumption. We empirically and theoretically argue against those two points and demonstrate that on any high-dimensional ($>$100) dataset, interpolation almost surely never happens. Those results challenge the validity of our current interpolation/extrapolation definition as an indicator of generalization performances.
\end{abstract}

\section{Introduction}

The origin of the interpolation and extrapolation notions are hard to trace back. \citet{kolmogoroff1941interpolation,Wiener1949ExtrapolationIA} defined extrapolation as predicting the future (realization) of a stationary Gaussian process based on past and current realizations. Conversely, interpolation was defined as predicting the possible realization of such process at a time position lying in-between observations, i.e., interpolation resamples the past. Various research communities have formalized those definitions as follows.

\begin{defn}
\label{def:interpolation}
Interpolation occurs for a sample $\bx$ whenever this sample belongs to the convex hull of a set of samples $\bX\triangleq \{\bx_1,\dots,\bx_{N}\}$, if not, extrapolation occurs.
\end{defn}

From the above definition, it is reasonable to assume extrapolation as being a more intricate task than interpolation. After all, interpolation guarantees that the sample lies within the dataset's convex hull, while extrapolation leaves the entire remaining space as a valid sample position.
Those terms have been ported {\em as-is} to various fields such as function approximation \citep{devore1998nonlinear} or machine learning \citep{bishop2006pattern}, and an increasing amount of research papers in deep learning provide results and intuitions relying on data interpolation \citep{belkin2018understand,bietti2019inductive,adlam2020neural}. Beyond those, the following adage ``as an algorithm transitions from interpolation to extrapolation, as its performance decreases'' is commonly agreed upon. Before going further, we insist that throughout this manuscript, interpolation is to be understood as characterizing the data geometry as per Def.~\ref{def:interpolation}. This is not to be mistaken with the often employed ``interpolation regime'' of models which occur whenever the latter has $0$ training loss on the data \citep{chatterji2021does}. We shall see that interpolation/extrapolation and generalization performances do not seem as tightly related as previously thought.

Our goal in this paper is to demonstrate both theoretically and empirically for both synthetic and real data that {\bf interpolation almost surely never occurs in high-dimensional spaces ($>100$) regardless of the underlying intrinsic dimension of the data manifold}. That is, given the realistic amount of data that can be carried by current computational capacities, it is extremely unlikely that a newly observed sample lies in the convex hull of that dataset. Hence, we claim that \setlist{nolistsep}
    \begin{itemize}[noitemsep]
    \item currently employed/deployed models are extrapolating
    \item given the super-human performances achieved by those models, extrapolation regime is not necessarily to be avoided, and is not an indicator of generalization performances
\end{itemize}
This paper is organized as follows. We first provide below (Thm.~\ref{thm:exponential}) an important theoretical result that has been derived in the context of Uniform samples from an hyper-ball. In that case, the probability of a new sample to be in interpolation regime from a dataset goes to $0$ as the dimension $d$ increases unless the number of dataset samples grows exponentially with $d$. This will allow to introduce notations and intuitions. We then directly provide empirical evidences in Sec.~\ref{sec:practice} using standard dataset where we demonstrate that even when considering a subset of the data dimensions, the probability to interpolate goes exponentially quickly to $0$ with the number of considered dimensions. We conclude with Sec.~\ref{sec:theory} by providing existing theoretical results describing the probability that new samples are in interpolation or extrapolation regimes in more specific scenarios.

\begin{thm}[\citet{barany1988shape}]
\label{thm:exponential}
Given a $d$-dimensional dataset $\bX\triangleq \{\bx_1,\dots,\bx_N \}$ with i.i.d. samples uniformly drawn from an hyperball, the probability that a new sample $\bx$ is in interpolation regime (recall Def.~\ref{def:interpolation}) has the following asymptotic behavior
$$
\lim_{d\rightarrow \infty} p(\underbrace{\bx \in \Hull(\bX)}_{\text{interpolation}}) = \begin{cases}
1 \iff N >d^{-1} 2^{d/2}\\
0\iff N <d^{-1} 2^{d/2}\\
\end{cases}
$$
\end{thm}

\section{Interpolation is Doomed by the Curse of Dimensionality}
\label{sec:practice}

In this section we propose various experiments supporting the need for exponentially large dataset to maintain interpolation, as per Thm.~\ref{thm:exponential}, for non Gaussian data. First, we demonstrate in Sec.~\ref{sec:intrinsic} the role of the underlying data manifold intrinsic dimension along with the role of the dimension of the smallest affine subspace that include the data manifold. As we will see from carefully designed datasets, only the latter has an impact on the probability of new samples being in an interpolation regime. We then move to real datasets in Sec.~\ref{sec:real_dataset} and demonstrate that both in the data space or in various embedding spaces, current test set samples are all in extrapolation regime from their corresponding training set.

\subsection{The Role of the Intrinsic, Ambient and Convex Hull Dimensions}
\label{sec:intrinsic}

\begin{figure}[t!]
    \centering
    \begin{minipage}{0.02\linewidth}
    \rotatebox{90}{$p(\bx \in \Hull(\bX))$}
    \end{minipage}
    \begin{minipage}{0.3\linewidth}
    \centering
        {\small $d^*=d$}\\
        \includegraphics[width=\linewidth]{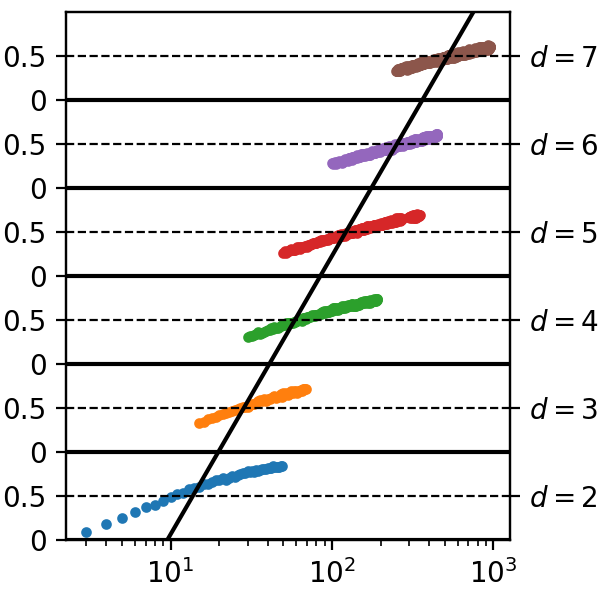}\\[-0.5em]
        \hspace{-0.3cm}$\log(N)$
    \end{minipage}
    \hfill
    \begin{minipage}{0.3\linewidth}
    \centering
        {\small $d^*=1$, nonlinear}\\
        \includegraphics[width=\linewidth]{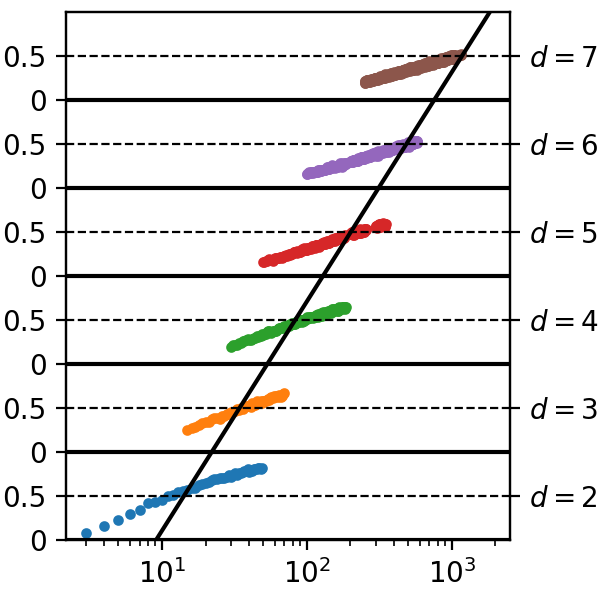}\\[-0.5em]
        \hspace{-0.3cm}$\log(N)$
    \end{minipage}
    \hfill
    \begin{minipage}{0.3\linewidth}
    \centering
        {\small $d^*=4$, linear}\\
        \includegraphics[width=\linewidth]{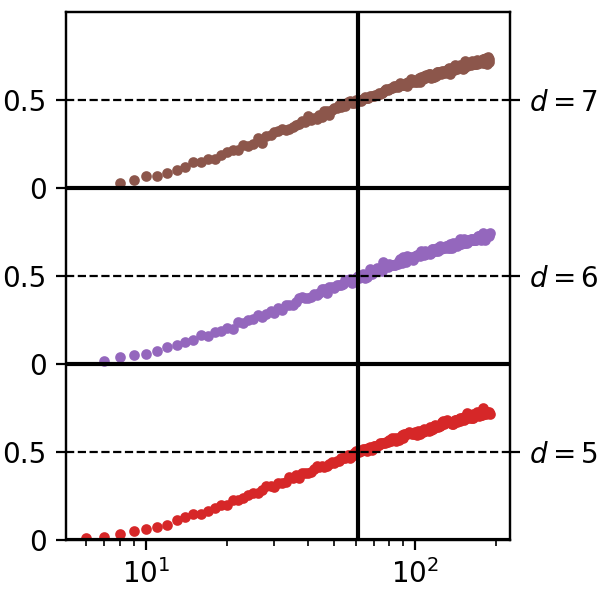}\\[-0.5em]
        \hspace{-0.3cm}$\log(N)$
    \end{minipage}
    \vspace{-0.2cm}
    \caption{\small Depiction of the evolution of the probability that a new sample is in interpolation regime (y-axis, $p(\bx \in \Hull(\bX))$) given increasing dataset size (x-axis, $N$) seen in logarithmic scale, and for various ambient space dimensions ($d$) based on Monte-Carlo estimates on $500,000$ trials. On the {\bf left}, the data is sampled from a Gaussian density $\bx_i\sim \mathcal{N}(0,I_d)$ while in the {\bf middle}, the data is sampled from a nonlinear continuous manifold with intrinsic dimension of $1$ (see Fig.~\ref{fig:manifold_data} for details on the manifold data) and on the {\bf right}, the data is sampled from a Gaussian density that lives in an affine subspace of constant dimension $4$ (while the ambient dimension increases). It is clear from those figures that {\bf in order to maintain a constant probability to be in interpolation regime, the training set size has to increase exponentially with $d^*$ regardless of the underlying intrinsic manifold dimension} where $d^*$ is the dimension of the lowest dimensional affine subspace including the entire data manifold i.e. the convex hull dimension.}
    \label{fig:evolution_toy}
\end{figure}

The first stage of our study consists in carefully understanding not only the role of the ambient dimension i.e. the dimension of the space in which the data lives, but also the role of the underlying data manifold intrinsic dimension i.e. the number of variables needed in a minimal representation of the data \citep{bennett1965representation}, and the dimension of the smallest affine subspace that includes all the data manifold.

In fact, one could argue that data such as images might lie on a low dimensional manifold and thus hope that interpolation occurs regardless of the high-dimensional ambient space. As we demonstrate in Fig.~\ref{fig:evolution_toy}, this intuition would be misleading. In fact, the underlying manifold dimension does not help even in the extreme case of having a $1$-dimensional manifold. What matters however, is the dimension $d^*$ of the smallest affine subspace that includes all the data manifold, or equivalently, the dimension of the convex hull of the data. As such, in the presence of a nonlinear manifold, we can see that the exponential requirement from Thm.~\ref{thm:exponential} in the number of samples required to preserve a constant probability to be in interpolation grows exponentially with $d^*$. In fact, with the intrinsic dimension ($d^*$) constant, increasing the ambient space dimension ($d$) has no impact on the number of samples needed to maintain interpolation regime as can be seen on the right of Fig.~\ref{fig:evolution_toy}. We thus conclude that {\bf for one to increase the probability to be in an interpolation regime, one should control $d^*$, and not the manifold underlying dimension not the ambient space dimension}.

\begin{figure}[t!]
    \centering
    \begin{minipage}{0.02\linewidth}
    \rotatebox{90}{\small dimension index \hspace{1.3cm}$z$\hspace{0.6cm}}
    \end{minipage}
    \begin{minipage}{0.97\linewidth}
    \centering
    \includegraphics[width=\linewidth]{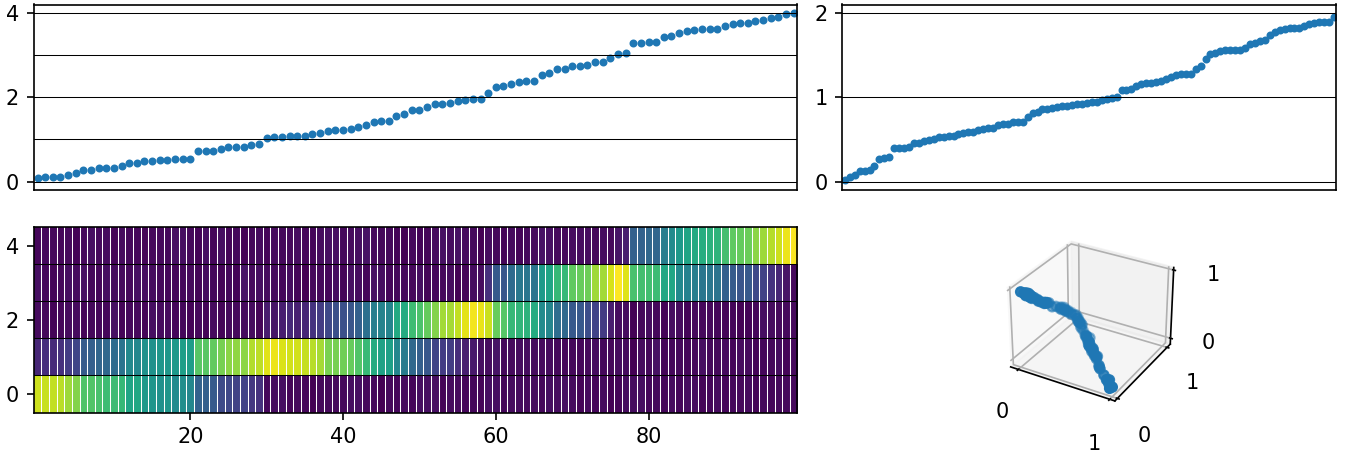}\\
    \vspace{-0.2cm}
    \hspace{-6cm}$N$
    \end{minipage}
    \vspace{-0.3cm}
    \caption{\small Depiction of the manifold data samples used for the middle plot of Figure.~\ref{fig:evolution_toy} with ${\rm dim}=5$ on the {\bf left} and ${\rm dim}=3$ on the {\bf right}. In all cases, the intrinsic dimension of this dataset is $1$, the latent coordinate ($z$) that governs the data ($\bx(z)$) is depicted on the {\bf top row} while the manifold samples in the ambient space are depicted in the {\bf bottom row}. This manifold is continuous, nonlinear and piecewise smooth, and corresponds to walking around the simplex.}
    \label{fig:manifold_data}
\end{figure}

We now propose to extend those insights to real data where the exact same behavior occurs across datasets.

\subsection{Real Datasets and Embeddings are no Exception}
\label{sec:real_dataset}

The previous section explored the cases of synthetic data with varying ambient, intrinsic, and convex hull dimensions. This provided valuable insights e.g. the key quantity of interest lies in the dimension of the smallest affine subspace containing the data. For real dataset however, one could argue that some natural properties of such manifolds help in being in an interpolation regime. Furthermore, one could argue that once embedded in a suitable and non-degenerate latent space, e.g. from a learned deep network, interpolation occurs. As we will see through various experiments, even with real datasets and various popular embeddings, interpolation remains an elusive goal that becomes exponentially difficult to reach as the dimension grows.

\begin{figure}[t!]
    \centering
    \begin{minipage}{0.02\linewidth}
        \rotatebox{90}{Proportion of test set in interpolation regime}
    \end{minipage}
    \begin{minipage}{0.97\linewidth}
    \centering
        \includegraphics[width=\linewidth]{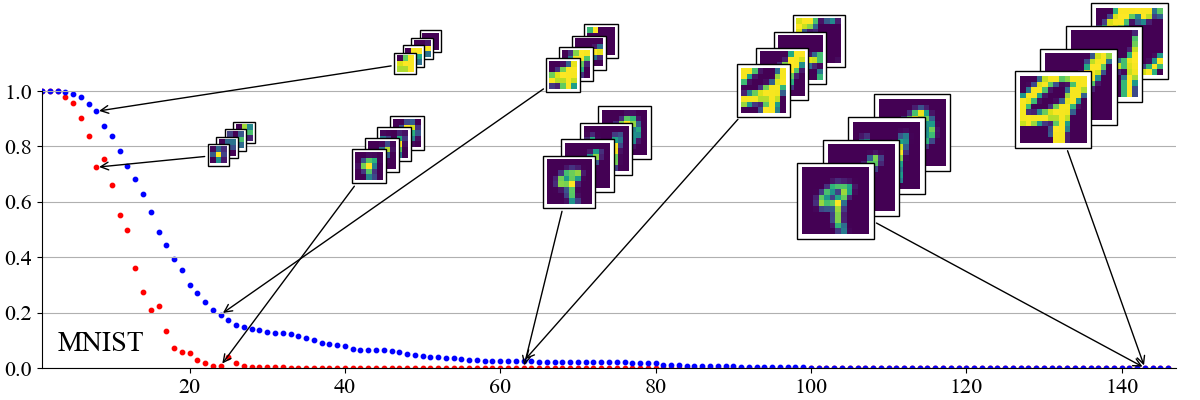}\\
        \includegraphics[width=\linewidth]{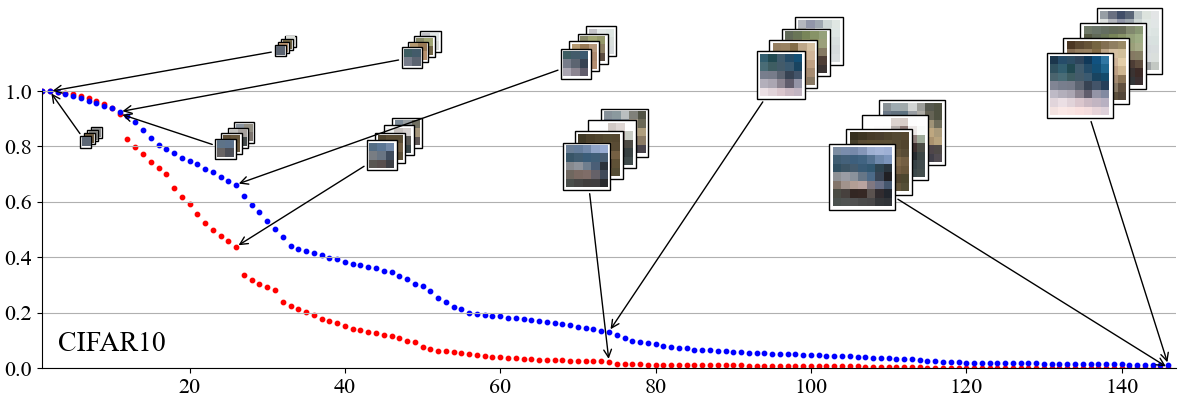}\\
        \includegraphics[width=\linewidth]{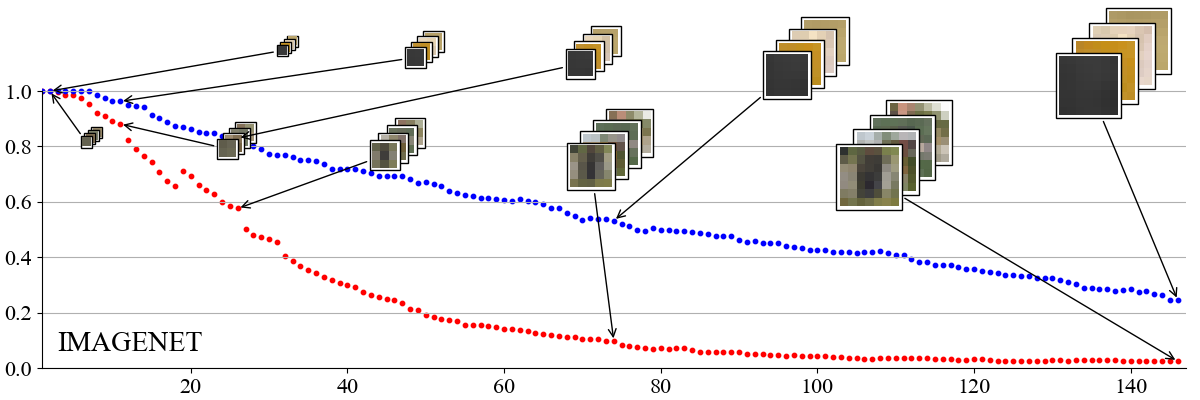}
    \end{minipage}
    considered number of dimensions $(d)$
    \vspace{-0.2cm}
    \caption{\small Depiction of the proportion of the test set that is in interpolation of the training set for MNIST ({\bf top}), CIFAR ({\bf middle}) and Imagenet ({\bf bottom}) as a function of the number of selected dimensions. We propose two settings ({\bf blue}) selecting increasingly large central patches (some cases consist of irregular patches for intermediate dimension values) and ({\bf red}) smoothing-subsampling the original images (some cases consist of irregular images for intermediate dimension values). Note that the blue line is always decreasing with $d$, and that $d=147$ (right of the x-axis) represents 19\% of MNIST total number of dimensions, 5\% for CIFAR and less than 1\% for Imagenet. As can be seen throughout those settings {\bf the proportion of the test set that is in interpolation regime decreases exponentially fast with respect to the number of dimensions} ultimately becoming negligible well prior reaching the full data dimensionality. The different slopes of those curves can be explained by the smallest dimensional affine space containing each type of data of (see Tab.~\ref{tab:test_set}).}
    \label{fig:evolution}
\end{figure}

\begin{figure}[t]
    \centering
    \begin{minipage}{0.01\linewidth}
    \rotatebox{90}{\small considered number of principal components}
    \end{minipage}
    \begin{minipage}{0.57\linewidth}
    \centering
    {\small\hspace{0.8cm} MNIST\hspace{1.6cm}CIFAR10\hspace{1.25cm}IMAGENET}\hspace{0.1cm}\\
    \includegraphics[width=1\linewidth]{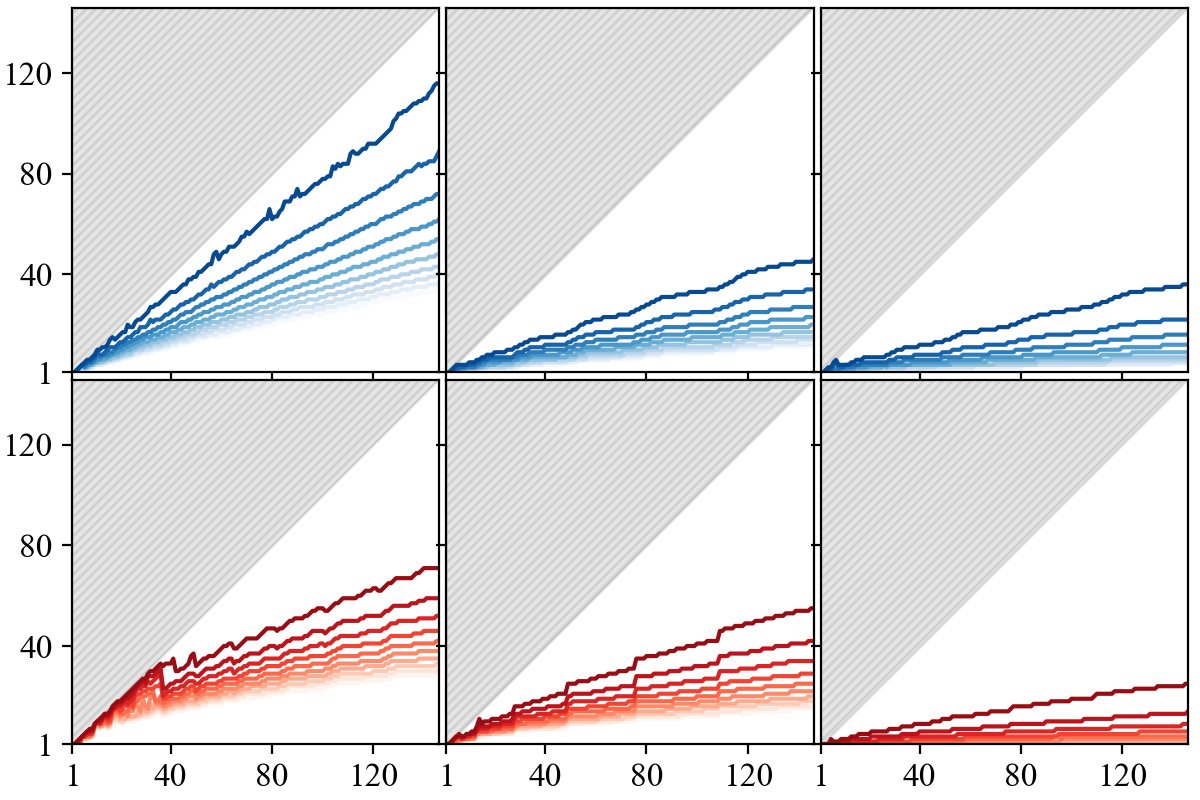}
    {\small considered number of sub-image dimensions (d)}
    \end{minipage}
    \begin{minipage}{0.39\linewidth}
    \caption{\small Depiction of levels ($90\%$ to $99\%$ from {\bf light} to {\bf dark}) of explained variance from a Principal Component Analysis model for varying sub-images dimensions ($1$ to $147$) {\bf x-axis} based on the number of considered components ({\bf y-axis}). The sub-images of dimension $d$ are obtained either by selecting the central spatial dimensions ({\bf blue, top-row}) or by smoothing and subsampling  ({\bf red, bottom-row}) as per Fig.~\ref{fig:evolution}. From this, it is clear that for each sub-image dimension ($d$), {\bf the smallest dimensional affine subspace containing the data reduces when going from MNIST to CIFAR10 to IMAGENET leading to the different slopes observed in Fig.~\ref{fig:evolution}} (recall Fig.~\ref{fig:evolution_toy}).}
    \label{fig:evolution_eigen}
    \end{minipage}
\end{figure}

{\bf Test set extrapolation in pixel-space.}~We first propose in Fig.~\ref{fig:evolution} to study the proportion of the test set that is in interpolation regime from the train set for MNIST, CIFAR and Imagenet. To grasp the impact of the dimensionality of the data we propose to compute this proportion with varying number of dimensions obtained from two strategies. First, we only keep a specified amount of dimensions from the center of the images, second, we smooth and subsample the images. The former has the benefit of preserving the manifold geometry whilst only considering a limited amount of dimensions, the latter preserves the overall geometry of the manifold while removing the high-frequency structures (details of the image) and compressing the information on fewer dimensions. In both cases and throughout datasets, we see that {\bf despite the data manifold geometry held by natural images, finding samples in interpolation regime becomes exponentially difficult with respect to the considered data dimension}.

{\bf Test set extrapolation in embedding-space.}~Given the above, one could argue that the key interest of machine learning is not to perform interpolation in the data space, but rather in a (learned) latent space. In fact, a DN provides a data embedding, then, in that space, a linear classifier (for example) solves the problem at hand, possibly in an interpolation regime. We thus provide in Tab.~\ref{tab:test_set} the proportion of the test set that is in interpolation regime when considering different embedding spaces. We observed that {\bf embedding-spaces provide seemingly organized representations (with linear separability of the classes), yet, interpolation remains an elusive goal even for embedding-spaces of only $30$ dimensions}. Hence current deep learning methods operate almost surely in an extrapolation regime in both the data space, and their embedding space.

\begin{figure}[t!]
    \centering
    \includegraphics[width=0.49\linewidth]{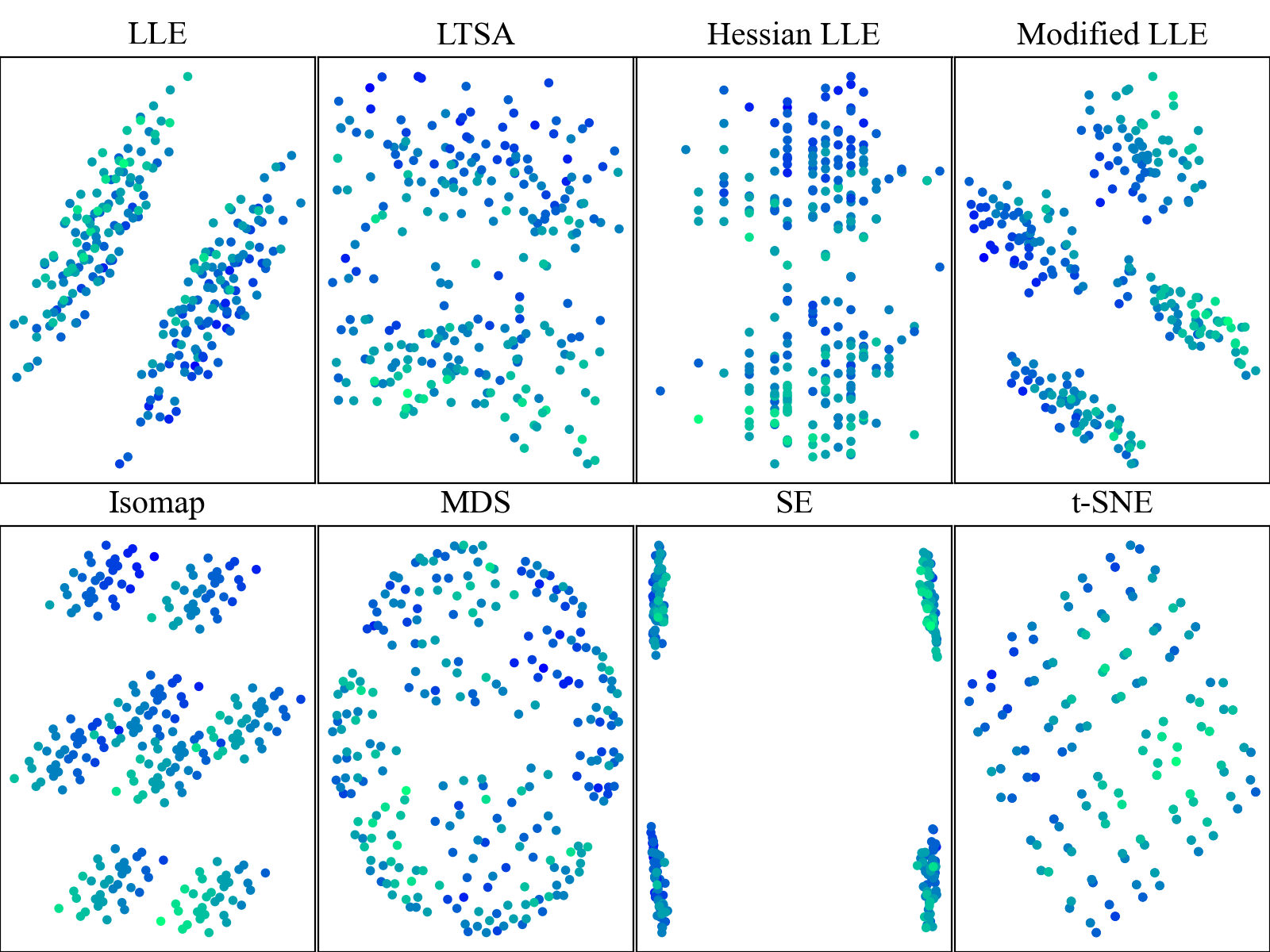}
    \includegraphics[width=0.49\linewidth]{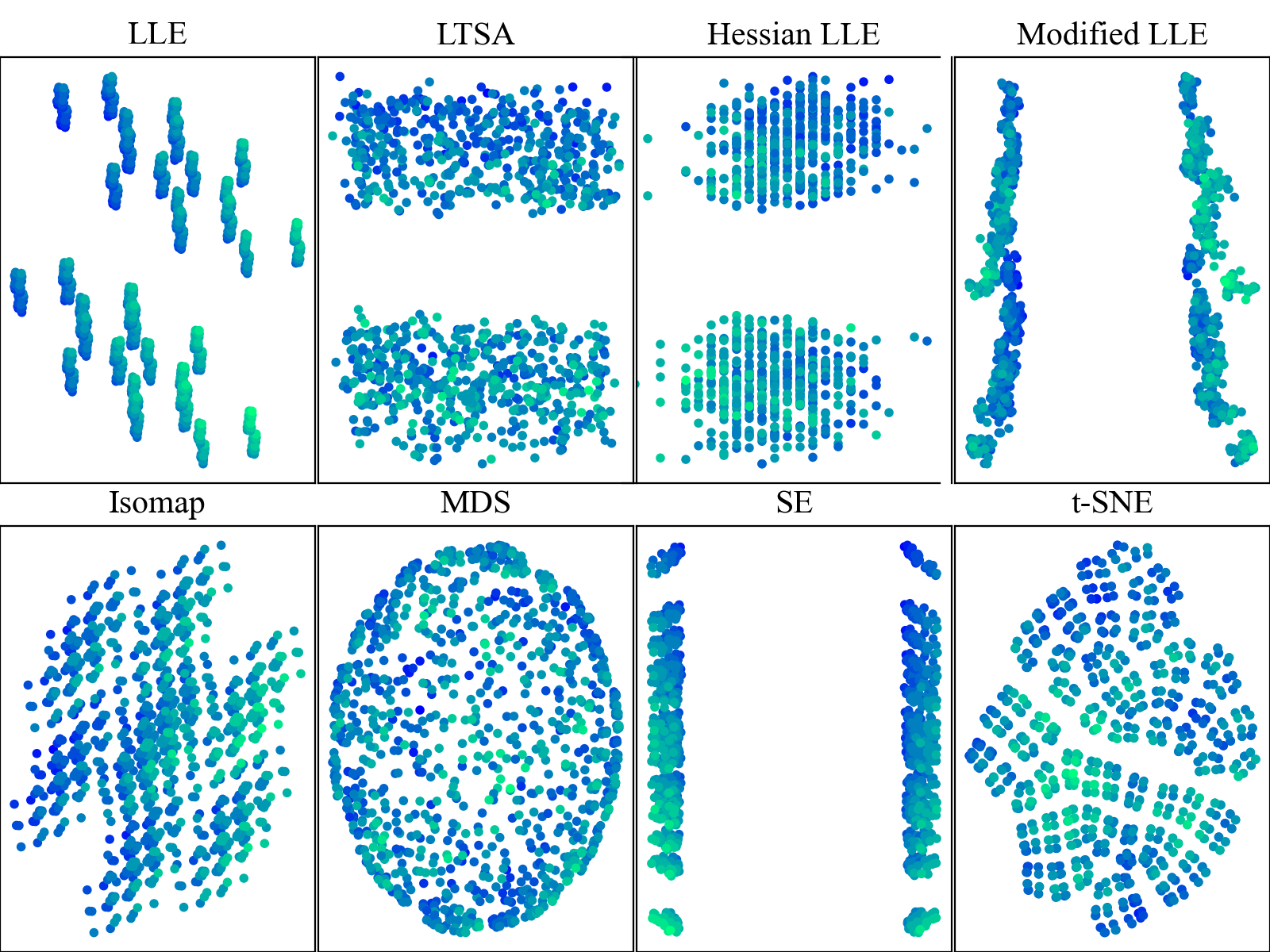}
    \caption{\small Depiction of various nonlinear dimensionality reduction techniques applied onto a synthetic dataset containing all the hypercube vertices for an hypercube of dimension $8$ ({\bf left}) and $10$ ({\bf right}). Coloring goes from blue for the vertex at position $(1,\dots,1)$ to green for the vertex at position $(-1,\dots,-1)$ in a linear manner. This data is chosen since {\bf each point/vertex is in extrapolation regime from all the other points/vertices}. However, existing techniques for dimensionality reduction primarily focus on preserving local geometric information. As a result, {\bf regardless of the employed dimensionality reduction algorithm, the  interpolation/extrapolation information is lost} as can be seen in all the proposed subplots. This can lead to hazardous assumptions and conclusions.}
    \label{fig:dim_reduction}
\end{figure}

{\bf Test set extrapolation in dimensionality-reduction-space.}~The last set of experiments deals with the use of (non)linear dimensionality reduction techniques to visualize high-dimensional dataset. We pose the following question: {\em is the interpolation/extrapolation information preserved by commonly employed dimensionality reduction techniques?} To unequivocally answer this question, we create a data that consists of the $2^d$ vertices of an hypercube in $d$ dimensions for $d=8,12$. Those dataset have the specificity that any sample is in extrapolation regime with respect to the other samples. We propose in Fig.~\ref{fig:dim_reduction} the $2$-dimensional representations of those vertices using $8$ different popular dimensionality reduction techniques: 
locally linear embedding \citep{roweis2000nonlinear} denoted as LLE, 
modified LLE \citep{zhang2007mlle}, 
Hessian eigenmaps \citep{donoho2003hessian} denoted as Hessian LLE, 
Laplacian eigenmaps \citep{belkin2003laplacian} denoted as SE, 
isomap \citep{balasubramanian2002isomap}, 
t-distributed stochastic neighbor embedding \citep{van2008visualizing} denoted as t-SNE,
local tangent space alignment \citep{zhang2004principal} referred as LTSA,
Multidimensional scaling \citep{kruskal1964multidimensional} denoted as MDS. We observe that {\bf dimensionality reduction methods loose the interpolation/extrapolation information and lead to visual misconceptions significantly skewed towards interpolation}.

{\bf Johnson–Lindenstrauss (di)lemma}
One last important setting concerns dimensionality reduction techniques that preserve -to some extent- the pairwise distances of the samples. Such techniques are often coined low-distortion embeddings, one of which follows from the Johnson–Lindenstrauss lemma (JLL) \citep{johnson1984extensions}.
In short, the JLL guarantees the existence of a linear mapping $f$ with input dimension $d$ and output dimension $d_{\rm JLL}\geq \frac{24}{3\epsilon ^2 - 2 \epsilon ^3} \log(N)$ with $N$ the size of the dataset such that the pairwise distances after projection will be within a $1\pm \epsilon$ factor of the original distances. Different bounds have emerged based on different proofs of the JLL (see \citet{dasgupta2003elementary} for a survey). Interestingly, as per Thm.~\ref{thm:exponential}, the experiments from Fig.~\ref{fig:evolution_toy} and Fig.~\ref{fig:evolution}, the dataset size must be of order $2^d$ to ensure that samples in the test set be in interpolation regime. In the JLL setting, this translates into $d_{\rm JLL}>\frac{24}{3\epsilon ^2 - 2 \epsilon ^3}d>d$. In other words, {\bf if a dataset size $N$ is exponential with the dimension $d$ (required to have new samples in interpolation regime) then $d_{\rm JLL}>d$ and JLL does not provide any dimensionality reduction}.

We propose in the next section for the interested reader a brief collection of theoretical results that have also reached the conclusion that in high-dimensional spaces, exponentially large datasets are required to maintain the probability for a new sample to be in interpolation regimes.

\begin{table}[t!]
    \centering
    \caption{\small Proportion (in $\%$) of the test set that falls into the interpolation regime for various datasets ({\bf rows}) and  embeddings ({\bf columns}) with varying (randomly selected) dimensions ($10$, $20$ and $30$), in all cases the embeddings are of dimension $512$. ``random projection'' represents a linear mapping with random Gaussian weights, the remaining six columns represent the use of the Resnet18 architecture's latent space with either untrained weights (usual initialization) or Imagenet-pretrained weights. We shall note that such results are not surprising based on the probabilities computed in Fig.~\ref{fig:evolution_toy} for synthetic data, since the embedded samples behave more closely to gaussian samples than in the original pixel-space.}
    \begin{tabular}{r|ccc|ccc|ccc}\cmidrule[2pt](){2-10}
         \multicolumn{1}{c|}{}
            &\multicolumn{3}{c|}{random projection}
            &\multicolumn{3}{c|}{random Resnet18}
            &\multicolumn{3}{c|}{pretrained Resnet18}
            \\\cmidrule[1pt](){1-10}
            \# selected dimensions & 10 & 20 & 30 & 10 & 20 & 30 & 10 & 20 & 30 \\\cmidrule[1pt](lr){1-10}
         \makecell{MNIST\\\footnotesize{$d$:764/train:50K/test:10K}} &83$\pm$1 & 12$\pm$2&0$\pm$0 & 94$\pm$3& 21$\pm$10& 0$\pm$0&95$\pm$2 & 52$\pm$17&14$\pm$12\\ 
         \makecell{CIFAR\\\footnotesize{$d$:3072/train:50K/test:10K}}  &89$\pm$1 &41$\pm$0 &15$\pm$0 & 88$\pm$4 & 23$\pm$8 &0$\pm$0 &92$\pm$2 &21$\pm$8 &0$\pm$0\\ 
         \makecell{Imagenet\\\footnotesize{$d$:150528/train:1M/test:100K}}  &81$\pm$4 &48$\pm$1 &22$\pm$1 & 91$\pm$1&30$\pm$1 &3$\pm$0 &90$\pm$2 &26$\pm$3 &1$\pm$0\\ \cmidrule[1pt](){1-10}
    \end{tabular}
    \label{tab:test_set}
\end{table}

\section{Theoretical Quantification of Interpolation Probabilities}
\label{sec:theory}

The previous section focuses on empirically evaluating the probability that a new sample falls into the convex hull of a given dataset and studied various settings concluding that interpolation suffers from the curse of dimensionality and is thus difficult to achieve in high-dimensional settings. As the convex hull is simply a polytope, we shall first refer the reader to \citet{spielman2004smoothed} for a thorough study between various properties of such polytopes and their relation to the data dimension. We then provide some milestone theoretical results on computing the probability that samples are in interpolation/extrapolation regime, a problem often named as the convex position problem.

\begin{defn}[Convex position problem]
For a convex body $K$ in the space, let $p(n,K)$ denote the probability that $n$ random, independent, and uniform points from $K$ are in convex position, that is, none of the samples lies in the convex hull of the others. 
\end{defn}

Note that with this formulation, $p(n,K)$ describes the probability that any point in the set of samples is in extrapolation regime.
Characterization of $p(n,K)$ for many $2$-dimensional bodies $K$ such as parallelograms and non-flat triangles are given below.

\begin{thm}[\cite{valtr1995probability,valtr1996probability}]
The probability $p(n,K)$ for a parallelogram or a triangle is given by 
\begin{align*}
    p(\underbrace{n,{\rm parallelogram}}_{\text{extrapolation}})=\left(\begin{pmatrix}2n-2\\n-1\end{pmatrix}/n!\right)^2\hspace{0.5cm}\text{ and }\hspace{0.5cm}
    p(\underbrace{n,{\rm triangle}}_{\text{extrapolation}})=\frac{2^n(3n-3)!}{(n-1)!^3(2n)!}.
\end{align*}
\end{thm}

Going to higher dimensional spaces has seen more challenging progresses making many existing results only valid in the limiting setting and when considering $K$ to be an hypersphere, as given in Thm.~\ref{thm:exponential} and in the following result.

\begin{thm}[\cite{buchta1986conjecture}]
The probability $p(n,K)$ for $K$ a $d$-dimensional hyperball $B^d$ and $n$ growing linearly with $d$ has the following limit behavior ,
$
\forall m > 3,  \lim_{d\rightarrow \infty}p(\underbrace{d+m,B^d}_{\text{extrapolation}}) = 1
$
\end{thm}

The above result effectively demonstrates that when sampling uniformly from an hyperball, the probability that all the points are in convex position (any sample lies outside of the other samples' convex hull) is $1$ when the number of samples grows linearly with the dimension, even when adding an arbitrary constant number of samples $m$. This result nicely complements the one provided in Thm.~\ref{thm:exponential} which was originally conjectured by \citet{buchta1986conjecture} a few years prior being proven by \citet{barany1988shape}.
More recently, a non asymptotic result has been obtained characterizing $p(n,K)$ when the samples are obtained from Gaussian distributions.

\begin{thm}[\cite{kabluchko2020absorption}]
Let $\bX$ consist of $N$ i.i.d. $d$-dimensional samples from $\mathcal{N}(0,I_d)$ with $N \geq d+1$, then for every $\sigma \geq 0$ the probability that a new sample $\bx\sim \mathcal{N}(0,\sigma^2Id)$ is in extrapolation regime is given by
\begin{align*}
    p(\underbrace{\bx \not \in \Hull(\bX)}_{\text{extrapolation}})=2(b_{N,d-1}(\sigma^2)+b_{N,d-3}(\sigma^2)+\dots)
\end{align*}
with
\begin{gather*}
    b_{n,k}(\sigma^2) = \begin{pmatrix}n\\k\end{pmatrix}g_k\left(-\frac{\sigma^2}{1+k\sigma^2}\right)g_{n-k}\left(\frac{\sigma^2}{1+k\sigma^2}\right),\;\;
    g_n(r) = \frac{1}{\sqrt{2\pi}}\int_{-\infty}^{\infty}\Phi^n\left(\sqrt{r}x\right)e^{-x^2/2}dx
\end{gather*}
where $\sqrt{r} = i\sqrt{-r}$ if $r<0$ and $b_{N,k}=0 $ for $k \not \in \{0,1,\dots,N\}$.
\end{thm}
The above theorem provides the analytical quantities governing the probability to be in interpolation regime. Lastly, \cite{majumdar2010random} consider the case where the samples are obtained from random walks and obtain similar interpolation behaviors. There remain many avenues to provide specific results when the data belong to lower dimensional manifolds, or to consider anisotropic distributions.

\section{Conclusion}

Interpolation and extrapolation, as per Def.~\ref{def:interpolation}, provide an intuitive geometrical characterization on the location of new samples with respect to a given dataset. Those terms are commonly used as geometrical proxy to predict a model's performances on unseen samples and many have reached the conclusion that a model's generalization performance depends on how a model interpolates. In other words, how accurate is a model within a dataset's convex-hull defines its generalization performances. In this paper, we proposed to debunk this (mis)conception. In particular, we opposed the use of interpolation and extrapolation as indicators of generalization performances by demonstrating both from existing theoretical results and from thorough experiments that, in order to maintain interpolation for new samples, the dataset size should grow exponentially with respect to the data dimension. In short, {\bf the behavior of a model within a training set's convex hull barely impacts that model's generalization performance since new samples lie almost surely outside of that convex hull}. This observation holds whether we are considering the original data space, or embeddings. We believe that those observations open the door to constructing better suited geometrical definitions of interpolation and extrapolation that align with generalization performances, especially in the context of high-dimensional data.

\bibliographystyle{apalike}
\bibliography{refs}

\end{document}